\documentclass[11pt]{article}%
\usepackage{amssymb}
\usepackage{float}
\usepackage{amsfonts}
\usepackage{amsmath}
\usepackage[nohead]{geometry}
\usepackage[doublespacing]{setspace}
\usepackage[bottom]{footmisc}
\usepackage{indentfirst}
\usepackage{graphicx}%
\usepackage{rotating}
\usepackage[english]{babel}
\usepackage{lineno}
\usepackage{natbib}
\usepackage{textcomp}
\usepackage{array}
\usepackage{gensymb}
\usepackage{empheq}
\usepackage[anythingbreaks]{breakurl}
\usepackage{mathtools}
\usepackage{lscape}
\usepackage{systeme}
\usepackage{booktabs}
\usepackage{multirow}

\usepackage{pdflscape}
\usepackage{dcolumn}

\usepackage{caption}
\usepackage{appendix}
\usepackage{subcaption}
\usepackage{hyperref}
\usepackage[T1]{fontenc}
\usepackage{booktabs} 
\usepackage{adjustbox,lipsum}
\usepackage{natbib}
\usepackage{footmisc}
\DefineFNsymbols{mySymbols}{{\ensuremath\dagger}{\ensuremath\ddagger}\S\P
   *{**}{\ensuremath{\dagger\dagger}}{\ensuremath{\ddagger\ddagger}}}
\setfnsymbol{mySymbols}
\setcitestyle{authoryear,open={},close={}}

\newcolumntype{L}{>{\centering\arraybackslash}m{3cm}}

\defcitealias{UN_SDG2016}{United Nations 2016}

\newcolumntype{C}[1]{>{\centering\arraybackslash}m{#1}}

\makeatletter
\def\@biblabel#1{\hspace*{-\labelsep}}
\makeatother
\begin{document}
\title{Predicting Illegal Fishing on the Patagonia Shelf from Oceanographic Seascapes}
\author{A. John Woodill, Maria Kavanaugh, Michael Harte, and James R. Watson\thanks{Correspondence: A. John Woodill, College of Earth, Ocean, and Atmospheric Sciences, Oregon State University, johnwoodill@gmail.com. We would like to thank Global Fish Watch for providing the vessel location data for our analysis. This work was supported by NASA under the award "Securing Sustainable Seas: Near real-time monitoring and predicting of global fishing fleet behavior" (Award No. 80NSSC19K0203). Additional support for developing seascapes was provided from NASA under the award " Dynamic Seascapes to Support a Biogeographic Framework for a Global Marine Biodiversity Observing Network " (Award No. 80NSSC18K0412). } \medskip\\  }
\maketitle

\singlespacing
\begin{center}
\textbf{Abstract}
\end{center}
\noindent Many of the world's most important fisheries are experiencing increases in illegal fishing, undermining efforts to sustainably conserve and manage fish stocks. A major challenge to ending illegal, unreported, and unregulated (IUU) fishing is improving our ability to identify whether a vessel is fishing illegally and where illegal fishing is likely to occur in the ocean. However, monitoring the oceans is costly, time-consuming, and logistically challenging for maritime authorities to patrol. To address this problem, we use vessel tracking data and machine learning to predict illegal fishing on the Patagonian Shelf, one of the world's most productive regions for fisheries. Specifically, we focus on Chinese fishing vessels, which have consistently fished illegally in this region. We combine vessel location data with oceanographic seascapes -- classes of oceanic areas based on oceanographic variables -- as well as other remotely sensed oceanographic variables to train a series of machine learning models of varying levels of complexity.  These models are able to predict whether a Chinese vessel is operating illegally with 69-96\% confidence, depending on the year and predictor variables used. These results offer a promising step towards preempting illegal activities, rather than reacting to them forensically.

\strut


\thispagestyle{empty}

\pagebreak%
\doublespacing

\newpage\section{INTRODUCTION} 


\noindent Illegal fishing adversely impacts the health of our oceans and will continue to challenge our attempts to achieve sustainable ocean use unless it is addressed. Recent estimates show illegal fishing represents as much as 20\% of the global fish catch and is valued at \$36.4 billion annually (\cite{sumaila2006, shaver2018}). These negative effects reach into local fishing communities that rely on sustainably managed fish stocks for sustenance, to support the local economy, and can even cause geopolitical instability that further undermines economic development (\cite{osterblom2012, mendenhall2020}). To address this important problem we must improve our ability to detect and predict illegal fishing (\cite{pramod2014, UN_SDG2016}). Illegal fishing undermines sustainability fisheries by reducing the accuracy of official fish catch and stock estimates, making it difficult for policymakers and regulators to manage local ecosystems (\cite{worm2012}). As a result, illegal fishing pushes total fishing effort beyond maximum sustainable levels thereby creating negative fish population growth rates and undermining our ability to monitor marine protected areas and sustain the health of the oceans (\cite{lubchenco2015}). 


One of the main challenges with the monitoring of illegal fishing is that the oceans are costly, time consuming, and logistically challenging for maritime authorities to patrol. This challenge becomes even more difficult because vessels will turn off their GPS transponders and go undetected or falsify their location -- this is known as "spoofing" (\cite{desouza2016}). Those vessels that spoof their location comprise a global "dark fleet" (\cite{sumaila2006}). To detect illegal fishing, recent research has focused on detecting vessels by detecting spoofing in their Automatic Identification System (AIS) data (\cite{ford2018, miller2018, petrossian2018}). However, while detection is key, it will only ever provide information about past events. Recent advances have targeted the prediction of illegal fishing, not just detection, by identifying anomalous spatial behaviors of fleets as they respond to nearby illegal activities (\cite{watson2019}). While the method can provide an early warning signal that illegal activities are soon to happen, precursors of illegal activities are unable to identify specific vessels as acting legally or not. An important next step is to improve our ability to predict in near real-time whether a fishing vessel is fishing legally or not, based solely on the ocean conditions of a given location.

Here, we demonstrate that illegal fishing can be predicted from oceanographic conditions, using the Patagonia Shelf as a case-study. The Patagonian Shelf is a region off the coast of Argentina that is known to be one of the most productive areas in the world for fisheries. Important marine ecosystems combined with year-round convective mixing and seasonal fronts provide abundant pelagic and demersal resources (\cite{csirke1987, arkhipkin2013}). This productive fishing region supports fishing trawlers, squid jiggers, and drifting longline fleets; these fleets have been extracting up to 1 million tonnes of squid annually, for example, consisting mostly of Argentinean and over 1 million tonnes of fish annually, including Hakes (\textit{Merluccius hubbsi}), Hoki (\textit{Macruronus magellanicus}), Rock Cod (Patagonotothen ramsayi) and Southern Blue Whiting (Micromesistius australis) (\cite{arkhipkin2013, laptikhovsky2013}). However, the region is heavily overfished. The UN Food and Agriculture Organization (UNFAO) provided a report in 2015 that showed 42\% of the Southwest Atlantic area was maintaining unsustainable stock levels, the third-worst among 15 major statistical areas (\cite{fao2018}). The report also highlights that the most important species (in terms of landings), Argentine shortfin squid (\textit{Illex argentinus}) and is "maximally sustainably fished." While the region provides productive ocean fishing conditions, overfishing combined with illegal fishing will have long term negative effects on the sustainability of the region.


Illegal fishing is rampant in the Patagonian Shelf. The former Argentine undersecretary of state for fisheries, Cesar Lerena, says the country is losing as much as \$5 billion US dollars and thousands of jobs annually (\cite{profeta2018}). Monitoring and enforcement in the region are difficult because legally and illegally operating fishing vessels operate alongside one another. Further, some of the most productive areas for fishing are located adjacent to the EEZ boundary. Therefore, fishing vessels can choose to operate legally just outside the EEZ or enter the EEZ to fish illegally, mostly undetected. Taking advantage of this spatial feature of the Patagonia shelf, Chinese fishing vessels have been shown to amass near Argentine waters at the EEZ in search of the shortfin squid (\cite{harkell2018}). Due to this, the sustainability of the squid stock is increasingly threatened as the difficulty to detect and predict illegal activity increases. 
		

In this paper, we have developed a method for predicting whether a given vessel is fishing illegally or not, based on its location and the oceanographic conditions it is experiencing. We focus on Chinese fishing vessels operating on the Patagonia shelf. China has the largest distant water fishing fleet of any nation and its vessels have engaged in numerous instances of illegal fishing in locations around the world (\cite{sala2018, profeta2018}). Importantly, Chinese fishing vessels do not have the right to fish in the Argentine Exclusive Economic Zone (EEZ). To identify whether a Chinese fishing vessel is fishing illegally in the Argentine EEZ, we developed a random forest classifier using oceanographic variables in combination with Automatic Identification System (AIS) vessel location tracking data. We then used the learning mechanism of the random forest model to identify the most important oceanographic predictors. Lastly, we also explored results when only an extremely limited set of oceanographic predictors were used  -- simulating a situation where only real-time data are available.  There is a growing emphasis on the use of machine learning to improve the detection and prediction of illegal fishing, and for quantifying total fishing effort in general (\cite{belhabib2019}).  A key advance in the use of machine learning is to use these methods to accurately distinguish legal from illegal fishing effort. Not only will this provide an estimate of a key yet hidden source of fish mortality (\cite{agnew2009}), critical for developing accurate and sustainable fisheries management policies, but this will provide a route to more efficient enforcement of the laws of the sea. 

\section{MATERIALS AND METHODS}

\subsection{Data}


\noindent Gridded AIS Vessel location data were obtained from Global Fish Watch (GFW) for the Patagonian Shelf region of South America for the period 2012-2016 (\cite{kroodsma2018}). This publically available data provide daily fishing vessel locations and fishing effort (in hours) at a 10th-degree spatial resolution, as well as a unique AIS identifier (MMSI). We also utilized vessel-specific characteristics by MMSI from GFW, such as flag (country) and gear type. For this analysis, we defined a vessel operating illegally as: (1) a Chinese fishing vessel (denoted by flag), (2) located inside the Argentine EEZ, with (3) positive fishing effort (i.e. in terms of the GFW data, this is identified in terms of non-zero fishing hours). In these data, the majority of illegal Chinese vessels were squid jiggers (81\%) and fishing trawlers (15\%) with most illegal fishing occurring in February and March (see Figure 2). Our dependent variable is imbalanced where the number of illegal fishing vessels makes up only 1.5\% of the total observed vessels. This subset provided 114,103 observations, 781 unique vessels operating legally, and 186 unique fishing vessels operating illegally (see Figure 1). 


Sea surface temperature (SST), chlorophyll (CHL), and seascape (SEA) observations were measured, using nearest neighbor interpolation, for each vessel location data point. Seascapes are a relatively new way to categorize ocean features in a way that preserves underlying data distributions, even when non-linear (\cite{ kavanaugh2016}). The marine environment can be viewed as a mosaic of distinct seascapes, with unique combinations of biological, chemical, geological, and physical processes that define habitats that change over time (See Table 1 for a list of Seascapes in the Patagonia Shelf; \cite{kavanaugh2014}). In addition to SST and CHL, which are typically two main oceanographic variables used to predict the spatial distribution of fishes (\cite{kleisner2010, roberts2016}), apex predators (\cite{scales2017}) and fishing (\cite{belhabib2019, crespo2018, soykan2014}), here we make use of seascapes to help summarize the numerous dimensions of variability in ocean conditions. SST and CHL are available from NASA Ocean Color and SEA are available from NOAA ERDDAP in 8-day increments (\cite{NASA2014, NASA2017, ERDDAP2016, kavanaugh2018}). Due to cloud cover, observations for every surface location were not available so we recorded this as "not available". We also used latitude and longitude coordinates of each vessel and whether they are inside the EEZ or not as predictor variables (see Table 2 for a list of all variables).  

\subsection{Predicting Illegal Fishing}


\noindent The overarching assumption with our data is that fishing vessels operate in locations where the ocean conditions are favorable for fishing. As such, vessels that are currently operating legally will evaluate their risk to operate illegally if ocean conditions change in such a way that incentivizes movement (\cite{watson2018}) to regions where they are not allowed to operate (i.e. inside the Argentine EEZ). When these conditions change and vessels choose to operate illegally, they receive a benefit from the change in ocean conditions but at a cost (risk) of getting caught through local enforcement (\cite{sumaila2006}). Our definition of what constitutes illegal fishing -- i.e. any Chinese vessel fishing in the Argentine EEZ -- is strict because we do not know if a particular vessel has a license to operate in the Argentine EEZ. Due to data limitations,  it is not possible to know which  Chinese vessels have licenses to fish in the Argentine EEZ, but we do know that this is only a small fraction of the total number of the Chinese fleet operating in the region (if any). So while our assumption that Chinese fishing vessels operating inside the EEZ, are operating illegally, is strict it is also conservative. 


To predict whether a Chinese vessel is operating illegally we developed a random forest classifier where the binary choice is defined as operating legally ($y=0$) or illegally ($y=1$). The classifier was fit with explanatory variables discussed above as well as a dummy variable,  specifically the month of the year due to the seasonality of fishing in the region. Our modeling strategy is similar to species distribution modeling where (illegal) fishing effort is linked to environmental variables to predict species habitat preferences (\cite{elith2009, russo2011, watson2016, ofarrell2017}). We diverged from these standard models to develop a binary classifier that can predict whether the vessel operated illegally. Our approach allows us to target individual vessels and identify where and when illegal activity is likely to occur.


Random forest classification is popular because it can handle complex relationships between explanatory variables (e.g. multicollinearity, non-linearity) by "bagging" the data into different subsets across different trees. These subset trees reduce the correlation across the data, thus reducing the variance and improving predictions. Another benefit of a random forest classifier is that it identifies the importance of each variable in terms of their contribution to prediction. This allowed us to include a variety of variables, identify those most important, and reduce our model down to a relatively simple model while maintaining classification skill. This "kitchen sink" approach does not just provide a useful prediction tool, but it improves our understanding of the drivers of illegal fishing. 


To validate our model, we utilized a block cross-validation technique where the test set was a single year and the training set was the remaining years. This technique produced five-folds, so validation is made on each fold (year). The reason for validating our model in this way is to prevent overfitting our model due to spatial autocorrelation (\cite{roberts2017}). 
When the training and test sets contain locations close (spatially) to each other and from the same year (temporally), cross-validation will predict those observations correctly while being over optimistic about observations further away. As a result, the trained model will not generalize to out-of-sample (test set) predictions, thus leading to overconfidence in the performance of the model. Therefore, we utilize a block cross-validation technique to ensure observations in the training and test set contain a similar spatial and temporal structure and are independent.


Since our data is imbalanced (more legal vessels than illegal vessels), we are interested in how well our model can predict illegal fishing ($y=1$) and not legal fishing ($y=0$) so our accuracy metrics should address the imbalance. We used five metrics to test the accuracy of our model: (1) Precision and Recall; (2) F-Score; (3) Area Under the Curve (AUC); (4) Average Precision (AP); and (5) Average Recall (AR). Precision is the fraction of true illegal activity relative to the number of false hits of illegal activities (the lower the precision, the more false positives included in predictions). Recall is the fraction of true illegal activity relative to the number of false misses of illegal activity (the lower the recall, the more false negatives). Precision describes the ability to predict true positive values whereas recall describes the ability to predict false positives. We utilize an F-1 score as our primary accuracy metric because it reflects the accuracy of a classifier with imbalanced classes (\cite{chawla2009}). An F-1 score measures the accuracy of the test set by using the harmonic mean of precision and recall where 100\% equals an accurate prediction. We also consider additional metrics to evaluate model performance. Precision and recall can be plotted against each other over different thresholds of predicted probabilities within the random forest model. From this comparison, AUC is computed by integrating under the curve for each year to find the level of accuracy. Finally, AP and AR summarize the precision-recall plot by using the weighted mean of precision (AP) and recall (AR) for each class (legal/illegal) at each threshold. If AP or AR equals one then this indicates zero false positives or false negatives in the predictions. For each fold, we estimated each of these metrics to validate the accuracy of the model.


Our first step to modeling illegal vessels relies on the "kitchen sink" approach where we used a large number of variables as predictors. Next, we utilized the learning mechanism of a random forest to identify predictors that are most important to prediction accuracy. We then created a separate model that included only the top five most important variables (SST, CHL, I-EEZ, LAT, LON). This simple model allowed us to reduce the complexity of the model to a subset of predictors. In addition, we also created a model with only oceanographic variables (SST, CHL, SEA).  Together, these reduced models were used to identify the specific ocean conditions that are related to illegal fishing on the Patagonia shelf. These reduced models were also created as an attempt to develop a methodology that could be operationalized with real-time data feeds, for which there is only a limited number of variables relative to the full suite of variables in historical datasets. 

\section{RESULTS}


\noindent The main results for predicting illegal vessels with all variables are provided in Figure 3 and Table 3 column 1. Figure 3 provides a precision-recall plot where predicted probabilities are varied across different thresholds to test the sensitivity of the results. Table 3 column 1 reports the F-1 score for each year that reflects how well the model can accurately predict illegal Chinese fishing vessels. Results show that the random forest classification algorithm can identify instances of illegal fishing -- as defined as a Chinese fishing vessel in the Argentine EEZ with non-zero fishing effort -- with an accuracy between 69-89\% depending on the year (i.e. between 2012-2016). Additional accuracy metrics are provided in Table S1 where Average-Precision (AP) ranges from 0.79-0.96, Average-Recall (AR) ranges from 0.69-0.99, and area-under-curve (AUC) ranges from 0.79-0.96. These results show we can confidently predict whether a vessel is operating illegally using oceanic biophysical characteristics and measures of geospatial distance. 


Next, we used the random forest classifier to identify the best predictors in our model. Figure 4 Panel A provides a list of the top five predictors and importance (variable importance is scaled to one). The top five variables that are important for predictions are longitude, latitude, sea surface temperature, chlorophyll, and whether the vessel is inside the Argentine EEZ. Using this information, we next created a reduced model that includes only these five variables, to again detect illegal fishing. The results from this analysis are provided in Table 3 column 2, and Table S1 column 2. The F-1 score accuracy ranges from 69-96\%, AP values range from 0.75-0.99, AR values range from 0.76-0.97, and AUC values range from 0.75-0.99. These results highlight that, somewhat counterintuitively, the reduced model is more accurate compared to our first “kitchen-sink" model. 


We next used only oceanographic variables: Seascape (SEA), sea-surface temperature (SST) and chlorophyll (CHL), and the month of the year; simulating a real-time situation where only these variables are available, and critically when there is an absence of vessel location data (See Figure 4 Panel B for top five oceanographic predictors). While learning in the random forest model improved our prediction accuracy, we also wanted to know how changes in oceanographic conditions can predict illegal fishing. Table 3 column 3 and Table S1 column 3 provides the results from this analysis. This ocean-only model showed that even with only oceanic biophysical variables, we can still classify illegal fishing with an accuracy of between 71-89\%. AP values range from 0.67-0.85, AR values range from 0.79-0.97, and AUC values range from 0.71-0.88. While this model does not perform as well as the previous two models it can inform us of when the ocean conditions are ripe for illegal fishing. 


Our model can predict whether a Chinese vessel is operating illegally on the Patagonia shelf; it can also provide insight into the behavior of fishing vessels in the region. If Chinese vessels expect better conditions for fishing tomorrow, will they decide to enter the EEZ and fish illegally? The best seascape predictor from the random forest model was Seascape 14: "Temperate Bloom Upwelling" (\cite{kavanaugh2014}). The temperate bloom upwelling seascape is common to subpolar and temperate waters and indicates upwelling which is almost always associated with productive fisheries. To explore the behavior of fishing vessels we tracked their movement from operating legally to illegally and the change in temperate bloom upwelling seascape. We estimated the number of legal fishing vessels that are not located in temperate upwell blooming seascape, but then move to this seascape the next day to operate illegally. Figure 5 shows this transition for temperate bloom upwelling as well as the other seascape transitions. In total, 83 Chinese fishing vessels decided to enter the EEZ and begin fishing illegally when they noticed these favorable oceanographic/fishing conditions. The “Temperate Transition" seascape was the next important seascape, with 80 vessels transitioning to this seascape to operate illegally. The temperate transition zone is typified by temperatures ranging from ~11-15 C, moderate to high salinities (>34 psu), low mesotrophic conditions ([chl-a]  0.2 to 0.5 mg m-3 with positive sea level anomalies indicative of convergent zones (see Table 1 for a list of seascape). These results suggest fishing vessels adjust their risk preferences to move to favorable areas of the ocean, which can be characterized by specific seascapes, to begin operating illegally at the risk of being caught to capture more fishing effort when conditions are favorable.


Finally, to gain an understanding of the regional oceanographic conditions associated with illegal fishing, we explored the spatial behavior of Chinese fishing vessels on the Patagonia shelf by examining the spatial distribution of the temperate bloom upwelling seascape when: (1) a day when high levels of illegal activity occur (n=76); and (2) when low levels of illegal activity occur (n=3). Figure 6 provides a map of legal fishing vessels (black dots), illegal fishing vessels (red dots), and the spatial distribution of the Temperate Bloom Upwelling seascape (blue). Note that the area covered by the temperate bloom upwelling seascape when there is a high level of illegal activity (panel A), is less than when there were low levels of illegal activity (panel B). This result suggests that when as upwelling/bloom seascape conditions diminish, this constricts the available fishing area so that vessels are more likely to move into the Argentine EEZ and operate illegally. The comparison made in Figure 6 provides further (qualitative) evidence of spatial behavioral responses to changes in ocean conditions.

\section{DISCUSSION}


\noindent To identify when illegal fishing occurs on the Patagonia shelf, we developed a random forest classifier trained on vessel location data and oceanographic information. We focused on Chinese fishing vessels operating illegally inside the Argentine EEZ, and used the spatial distribution of different seascapes as predictors, amongst others. We performed cross-validation by individual years to measure the skill of our predictions and found that the model can predict illegal fishing with an accuracy of 69-89\%. We also created a reduced model, one with a restricted set of important predictor variables, and showed that the random forest classifier can still accurately predict with 69-96\% confidence. Further still, using oceanographic predictor variables only, we still achieved a level of accuracy from 71-89\%. Finally, we explored how fishers' spatial behaviors respond to changes in a specific seascape -- the temperate blooms upwelling seascape, which we find is the seascape to which fishers move to fish (illegally). A recent paper provided a new method for measuring illegal fishing effort based on the spatial behavior of vessels located in an illegal fishing area (\cite{belhabib2019}). Our method improves upon this method by providing the critical first step in classifying whether the fishing vessel is, in fact, operating illegally, based on oceanic conditions. These results provide important insights into the limits to our predictive skill and the (spatial) behavioral choices of fishing vessels that commit these crimes.


The methodology introduced here is a necessary first step to making (near) real-time predictions of illegal activities at sea, potentially for the whole globe. To make subsequent improvements, an important next step is to expand the scope of our analysis to include other important regions where illegal fishing occurs. One challenge will be that the Patagonian Shelf is a unique case study because the Argentine EEZ border essentially runs straight through a major fishing hotspot, leading to foreign fishing vessels essentially “fishing along the line" of the EEZ. As a consequence, foreign vessels are often tempted (by changing ocean conditions or by changing spatial distributions of target species) to make incursions into the Argentine EEZ. This specific spatial nature of the Patagonia shelf system will not be represented in other systems around the world, and how and why illegal fishing will occur will be nuanced in ways that new machine learning approaches might be required in other regions.


To improve the utility of our machine learning approach, it can (and will) be combined with other data and methods. For example, recent advances in the detection of anomalous multiscale patterns in fleet spatial behaviors have shown that (spatial) early-warning signals proceed  IUU activities ((\cite{watson2019}). Methods like these can be combined with the machine learning method presented here, making it possible to refine the situational awareness of IUU ongoing in a particular region. No single approach offers a panacea. Essentially, a suite of IUU detection and prediction techniques can (and should) be employed in concert to identify potential "hot spots" of illegal activity. Doing so will greatly increase the efficiency by which enforcement organizations monitor/patrol areas of our oceans. 


A major caveat to note is that we have defined illegal fishing specifically in terms of Chinese vessels operating in the Argentine EEZ, which we have assumed they have no right to do. In general, this should be true, as most Chinese vessels that do not have the rights to fish in Argentine waters (\cite{mallory2013}). But we cannot say with complete confidence that this is so. As a consequence, there is a risk that our analysis includes false positives. Further, our analysis shows that vessels (of all nations except Argentina) operate extremely close to the EEZ line. Vessels may not be aware of the exact location of the EEZ line, and perhaps even "a little" incursion into Argentine waters might be the norm. To explore the impact of the proximity of fishing vessels to the EEZ line, we repeated our analysis but with an increased buffer zone around the EEZ line; specifically, illegal fishing was redefined as a Chinese vessel fishing within 2, 5 and 10 kilometers (km) of the EEZ line. Table 3 columns 4-6 and Table S1 columns 4-6 provides the results of this sensitivity analysis. The 2km and 5km buffer results show the model is still able to predict illegal fishing with an F-1 accuracy score between 45-92\%. However, at a 10km buffer, the results are less stable across all accuracy metrics. In 2012, there were no fishing vessels operating illegally inside the EEZ so we are unable to evaluate accuracy metrics. For 2013-2016, the F-1 scores range between 0 and 92\%. While the model does break down with a buffer at 10km, it still performs rather well with a 2-5km buffer. This suggests that it is possible to predict the short-distance but frequent incursions into Argentine waters, but not the long-distance incursions that are rare. 


An important result from this analysis is the need for finer data resolution to predict illegal activity. While our model does well at predicting illegal activity when vessels incur  <5km into the EEZ, the model accuracy drops when vessel incursions are  >10km from the EEZ border. The important point to note is that the resolution of the oceanographic data used is 10km, meaning the model results are confined to this spatial scale also. Therefore, to make accurate determinations of illegal fishing,  higher spatial and temporal resolution oceanographic data is required. Higher spatial resolution  (300 m to 1 km) ocean color is available from moderate resolution spectrometers on  NASA and ESA satellites. For example, the proposed PACE (Phytoplankton Aerosol Carbon and Ecosystems) mission of NASA will provide hyperspectral remote sensing reflectances which will increase the capacity of imagery to provide information on phytoplankton size distribution and functional type, and thus bloom succession and lower trophic level food quality.  However, all polar orbiting satellites tend to have repeats on the order of 1-2 days, which may minimize the utility of a near real-time monitoring scheme, given travel times and speeds of industrialized fishing vessels. Geostationary satellites, however, provide near continuous monitoring of a region, potentially providing hourly information on the location of features. Currently, the only geostationary satellite with operational ocean color is the South Korean Geostationary Ocean Color Imager (GOCI) that provides 500 m resolution ocean color data for the Korean Sea and the northern East China Sea.

\section{CONCLUSION}


\noindent To conclude, deterring the impact of illegal fishing will improve the health and sustainability of our oceans. The impact of illegal fishing reaches into regional and local economies and even causes geopolitical tensions which further reduces the growth of developing countries' economies. Countries have begun improving maritime monitoring and law enforcement; however, recent advances rely on a retrospective analysis of individual vessel spoofing which is unable to provide real-time information. Going forward, it is vital that law enforcement be able to prevent illegal activities. To make steps in this direction, we have developed a machine learning algorithm that can accurately predict whether a Chinese fishing vessel is operating illegally with 69-96\% accuracy in the Patagonian Shelf region of Argentina. In addition to prediction, our use of machine learning has informed us of why illegal fishing happens in the region based on changing ocean conditions. These results offer a promising step towards real-time monitoring and likelihood estimates of "hot spots" where illegal fishing is likely to occur based on changes in oceanographic conditions. As new information and data become available, these valuable tools will allow maritime law enforcement to improve their ability to monitor and enforce illegal activity in the oceans, thus reducing illegal fishing and improving the sustainability of fish stocks.

\section{ACKNOWLEDGEMENTS}
\noindent Correspondence: A. John Woodill, Oregon State University, johnwoodill@gmail.com. We would like to thank Global Fish Watch for providing the vessel location data for our analysis. 
This work was supported by NASA under the award "Securing Sustainable Seas: Near real-time monitoring and predicting of global fishing fleet behavior" (Award No. 80NSSC19K0203). 
Additional support for developing seascapes was provided from NASA under the award " Dynamic Seascapes to Support a Biogeographic Framework for a Global Marine Biodiversity Observing Network " (Award No. 80NSSC18K0412)

\section{DATA AVAILABILITY}
\noindent The AIS data to support the findings in this study are available at Global Fish Watch (\url{https://globalfishingwatch.org/datasets-and-code/}). Oceanographic data are available at NASA Ocean Color (\url{https://oceancolor.gsfc.nasa.gov/}) and ERDDAP (\url{https://cwcgom.aoml.noaa.gov/erddap/griddap/noaa_aoml_seascapes_8day.graph}). Code to support the findings in this study are available online.
\newpage
\bibliographystyle{apalike}

\newcolumntype{P}[1]{>{\raggedright\arraybackslash}p{#1}}

\begin{table}[!htbp] \centering

\begin{tabular}{P{1.5cm}P{4cm}P{8cm}} 
\\[-1.8ex]\hline 
\hline \\[-1.8ex] 
Class & Name & Description\\ 
\hline \\[-1.8ex] 
1 & North Atlantic Spring, ACC Transition & Cold (4-7 C), low dynamic topography, moderate nutrient stress \\ 
2 & Subpolar-Subtropical  Transition & Variable SST (12-21C), Oligotrophic to mesotrophic, moderate to high nutrient stress \\ 
7 & Temperate Transition & Temperatures ranging from ~ 11-15 C, moderate to high salinities (>34 psu), low mesotrophic conditions ([chl-a]  0.2 to 0.5 mg m-3 with positive sea level anomalies indicative of convergent zones \\ 
12 & Subpolar & Cool (6-9 C), mesotrophic, low CDOM \\ 
14 & Cool Temperate Blooms Upwelling & Subpolar and temperate waters -- indicates upwelling which is almost always associated with productive fisheries \\ 
15 & Tropical Seas & Warm, salty (~35psu), low mesotrophic conditions \\ 
17 & Subtropical Transition (low nutrient stress) & Warm, mesotrophic, low nutrient stress \\ 
19 & Arctic/Freshwater influenced Subpolar Shelves & Lower salinity <32, high CDOM \\ 
21 & Warm, Blooms, High Nuts & 18-22 degrees C, high chl-a (>2 mg m-3), high CDOM, salinity variable but relatively fresh (32 psu) \\

\hline \\[-1.8ex] 
\end{tabular}
\parbox{5.75in}{\textbf{Table 1: Summary of Patagonian Shelf Seascapes} \\
Table provides a list of the most prevalent seascapes in the Patagonian Shelf region. }
\label{table:alt_spray_strats}
\end{table}

\newpage

\newcolumntype{P}[1]{>{\raggedright\arraybackslash}p{#1}}

\begin{table}[!htbp] \centering 

\begin{tabular}{P{2.25cm}P{1.25cm}P{6cm}P{6cm}}
\\[-1.8ex]\hline 
\hline \\[-1.8ex] 
Variable & Abbr. & Description & Source \\ 
\hline \\[-1.8ex] 
Sea Surface Temperature &
  SST &
  Skin sea surface temperature in units of Celcius. &
  NASA Ocean Color Level-3 MODIS-Aqua 8-day SST at 4km resolution. \\ 
Chlorophyll     & CHL   & Near-surface concentration of chlorophyll-a                                 & NASA Ocean Color Level-3 MODIS-Aqua 8-day CHL at 4km resolution. \\ 
Seascapes &
  SEA &
  Classification (n=30) of ocean features as seascapes derived from \newline CHL, SST, and photosynthetically available radiation (PAR). &
  NOAA ERDDAP 8-day SEA at 5km resolution. \\
Latitude        & LON   & Geographical coordinate of fishing vessel along the east-west axis.         & AIS Global Fishwatch Data                                        \\ 
Longitude       & LAT   & Geographical coordinate of fishing vessel along the north-south axis.       & AIS Global Fishwatch Data                                        \\ 
Inside EEZ      & I-EEZ & Binary variable describing whether fishing vessel is located inside the EEZ & AIS Global Fishwatch Data                                        \\ 
Distance to EEZ & D-EEZ & The distance a fishing vessel is from the EEZ (in kilometers)               & AIS Global Fishwatch Data                                        \\ 
\hline \\[-1.8ex] 
\end{tabular}
\parbox{6.75in}{\textbf{Table 2: Summary of Random Forest Variables} \\
Table provides a list of variables, abbreviations, a brief description and the source of the data used in the random forest classifier. }
\label{table:alt_spray_strats}
\end{table}

\newpage

\begin{table}[H]

\small

\begin{tabular}{|c|c|c|c|c|c|c|}
\hline
\multirow{2}{*}{} & \textbf{\begin{tabular}[c]{@{}c@{}}Random\\ Forest\end{tabular}} & \textbf{\begin{tabular}[c]{@{}c@{}}Random\\ Forest\end{tabular}} & \textbf{\begin{tabular}[c]{@{}c@{}}Random\\ Forest\end{tabular}} & \textbf{\begin{tabular}[c]{@{}c@{}}Random\\ Forest\end{tabular}} & \textbf{\begin{tabular}[c]{@{}c@{}}Random\\ Forest\end{tabular}} & \textbf{\begin{tabular}[c]{@{}c@{}}Random\\ Forest\end{tabular}} \\ \cline{2-7} 
 & \textit{\begin{tabular}[c]{@{}c@{}}all\\ variables\end{tabular}} & \textit{\begin{tabular}[c]{@{}c@{}}top five \\ variables\end{tabular}} & \textit{\begin{tabular}[c]{@{}c@{}}Seascape, \\ SST, CHL, \\ Month\end{tabular}} & \textit{\begin{tabular}[c]{@{}c@{}}all variables\\ 2km EEZ\end{tabular}} & \textit{\begin{tabular}[c]{@{}c@{}}all variables\\ 5km EEZ\end{tabular}} & \textit{\begin{tabular}[c]{@{}c@{}}all variables\\ 10km EEZ\end{tabular}} \\ \hline
 & (1) & (2) & (3) & (4) & (5) & (6) \\ \hline
\textbf{F-1 Score} & \multicolumn{6}{c|}{\textbf{}} \\ \hline
2012 & 69\% & 84\% & 89\% & 92\% & 80\% & NA \\ \hline
2013 & 70\% & 74\% & 77\% & 58\% & 47\% & 0 \\ \hline
2014 & 55\% & 69\% & 75\% & 79\% & 45\% & 0 \\ \hline
2015 & 89\% & 93\% & 71\% & 81\% & 78\% & 92\% \\ \hline
2016 & 80\% & 96\% & 73\% & 92\% & 40\% & 67\% \\ \hline
\end{tabular}
\parbox{6in}{\textbf{Table 3: Summary of Main Results} \\
Table provides results from our analysis across different model specifications shown in column: (1) using all variables; (2) top variables from random-forest; (3) Seascape and Distance measures; (4-6) sensitivity analysis of vessels distance to EEZ being inside 2km, 5km, and 10km. Table provides an F-1 Score to evaluate model performance for each predicted year 2012-2016. See Table S1 for additiona accuracy metrics.}
\end{table}

\newpage

\begin{table}[H]

\small

\begin{tabular}{|c|c|c|c|c|c|c|}
\hline
\multirow{2}{*}{} & \textbf{\begin{tabular}[c]{@{}c@{}}Random\\ Forest\end{tabular}} & \textbf{\begin{tabular}[c]{@{}c@{}}Random\\ Forest\end{tabular}} & \textbf{\begin{tabular}[c]{@{}c@{}}Random\\ Forest\end{tabular}} & \textbf{\begin{tabular}[c]{@{}c@{}}Random\\ Forest\end{tabular}} & \textbf{\begin{tabular}[c]{@{}c@{}}Random\\ Forest\end{tabular}} & \textbf{\begin{tabular}[c]{@{}c@{}}Random\\ Forest\end{tabular}} \\ \cline{2-7} 
 & \textit{\begin{tabular}[c]{@{}c@{}}all\\ variables\end{tabular}} & \textit{\begin{tabular}[c]{@{}c@{}}top five \\ variables\end{tabular}} & \textit{\begin{tabular}[c]{@{}c@{}}Seascape, \\ SST, CHL, \\ Month\end{tabular}} & \textit{\begin{tabular}[c]{@{}c@{}}all variables\\ 2km EEZ\end{tabular}} & \textit{\begin{tabular}[c]{@{}c@{}}all variables\\ 5km EEZ\end{tabular}} & \textit{\begin{tabular}[c]{@{}c@{}}all variables\\ 10km EEZ\end{tabular}} \\ \hline
 & (1) & (2) & (3) & (4) & (5) & (6) \\ \hline
\textbf{Average-Precision} & \multicolumn{6}{c|}{\textbf{}} \\ \hline
2012 & 0.90 & 0.80 & 0.85 & 0.96 & 0.82 & NA \\ \hline
2013 & 0.79 & 0.75 & 0.67 & 0.41 & 0.22 & 0 \\ \hline
2014 & 0.83 & 0.84 & 0.77 & 0.86 & 0.49 & 0.20 \\ \hline
2015 & 0.96 & 0.94 & 0.77 & 0.95 & 0.89 & 1 \\ \hline
2016 & 0.92 & 0.99 & 0.79 & 0.99 & 0.87 & 0.71 \\ \hline
\textbf{Average-Recall} & \multicolumn{6}{c|}{\textbf{}} \\ \hline
2012 & 0.99 & 0.97 & 0.97 & 0.99 & 0.99 & NA \\ \hline
2013 & 0.79 & 0.86 & 0.92 & 0.82 & 0.72 & 0.50 \\ \hline
2014 & 0.69 & 0.76 & 0.82 & 0.87 & 0.67 & 0.50 \\ \hline
2015 & 0.91 & 0.96 & 0.97 & 0.84 & 0.82 & 0.92 \\ \hline
2016 & 0.83 & 0.97 & 0.80 & 0.95 & 0.63 & 0.77 \\ \hline
\textbf{Area-Under-Curve} & \multicolumn{6}{c|}{\textbf{}} \\ \hline
2012 & 0.90 & 0.78 & 0.88 & 0.96 & 0.8 & NA \\ \hline
2013 & 0.79 & 0.75 & 0.71 & 0.46 & 0.18 & 0 \\ \hline
2014 & 0.83 & 0.85 & 0.76 & 0.87 & 0.48 & 0.21 \\ \hline
2015 & 0.96 & 0.94 & 0.78 & 0.95 & 0.89 & 1 \\ \hline
2016 & 0.92 & 0.99 & 0.79 & 0.99 & 0.87 & 0.71 \\ \hline

\end{tabular}
\parbox{6.5in}{\textbf{Table S1: Summary of Additional Results} \\
The table provides results from our analysis across different model specifications shown in column: (1) using all variables; (2) top variables from random-forest; (3) Seascape and Distance measures; (4-6) sensitivity analysis of vessels distance to EEZ being inside 2km, 5km, and 10km. The table provides Average-Precision, Average-Recall, and Area-Under-Curve accuracy metrics to evaluate model performance for each predicted year 2012-2016.}
\end{table}

\newpage

\captionsetup{labelfont=bf}

\begin{figure}[H]
    \centering
    \includegraphics[]{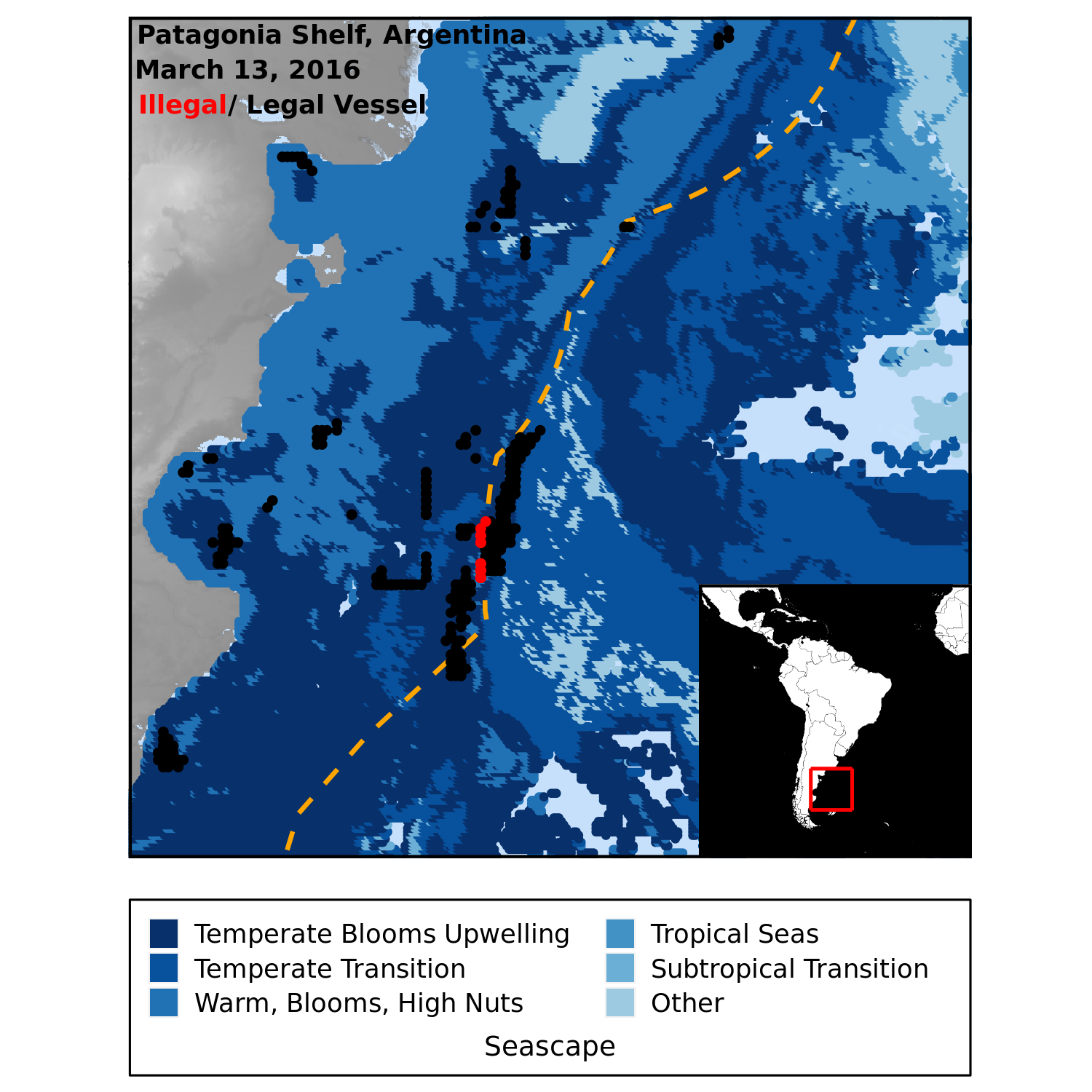}
        \caption[]{\textbf{Map of Patagonian Shelf with Seascapes and Illegal/Legal Fishing Vessels}{\\ 
        Map of the Patagonian shelf region with the EEZ (orange dashed line), five most prevalent seascapes (colored in blue), and the location of vessels on March 13, 2016. Chinese vessels fishing illegally, i.e. within the Argentine EEZ, are highlighted by red markers.  

}}
    \label{fig:figure1}
    \end{figure}

\newpage

\begin{figure}[H]
    \centering
    \includegraphics[]{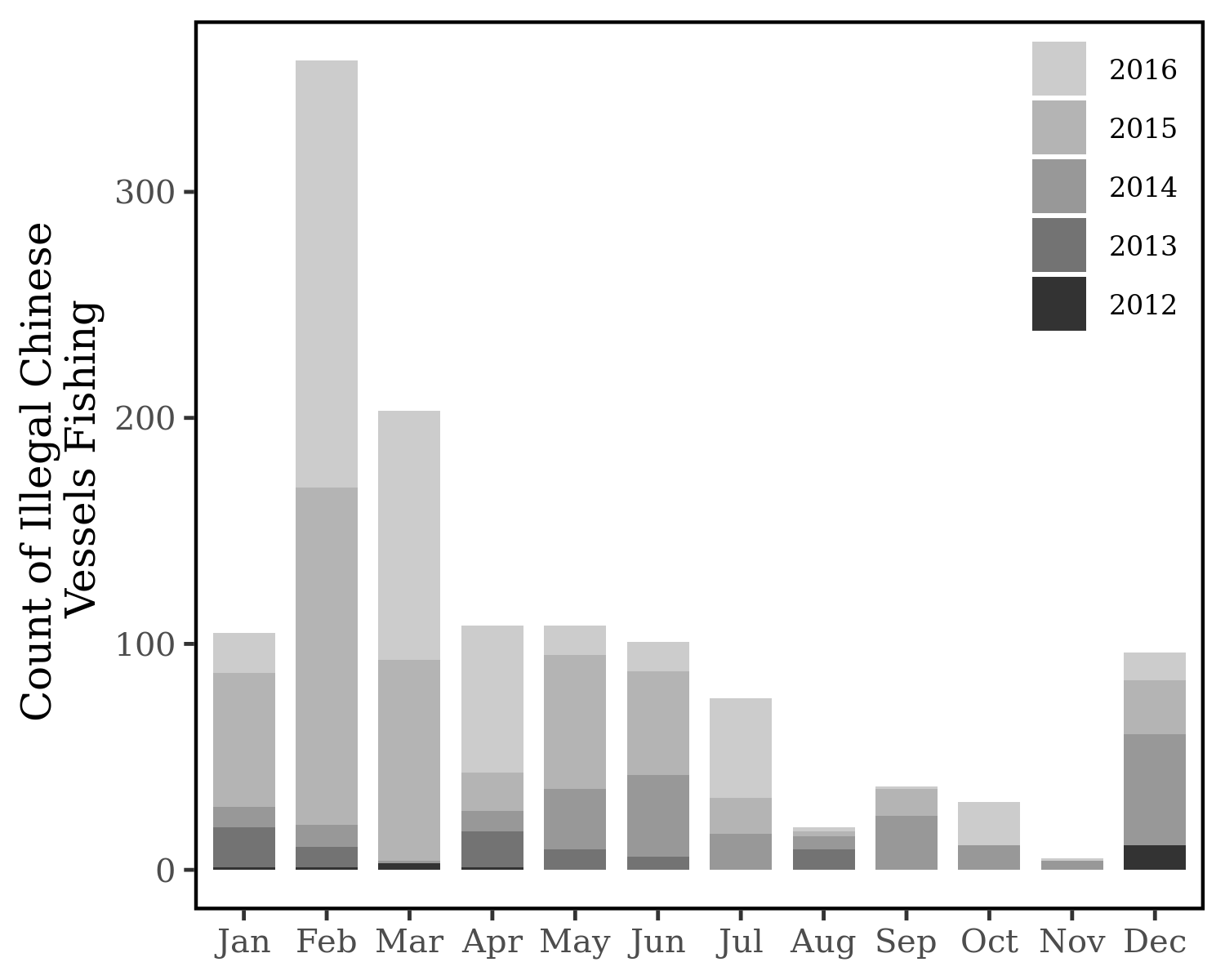}
        \caption[]{\textbf{Number of Illegal Events per Month and Year}{\\
        Figure provides the number of illegal events by month and year in the Patagonia Shelf region.
The majority of illegal Chinese fishing occurs in February and March when the squid fishery is most active.

}}
    \label{fig:figure2}
\end{figure}

\newpage

\begin{figure}[H]
    \centering
    \includegraphics[]{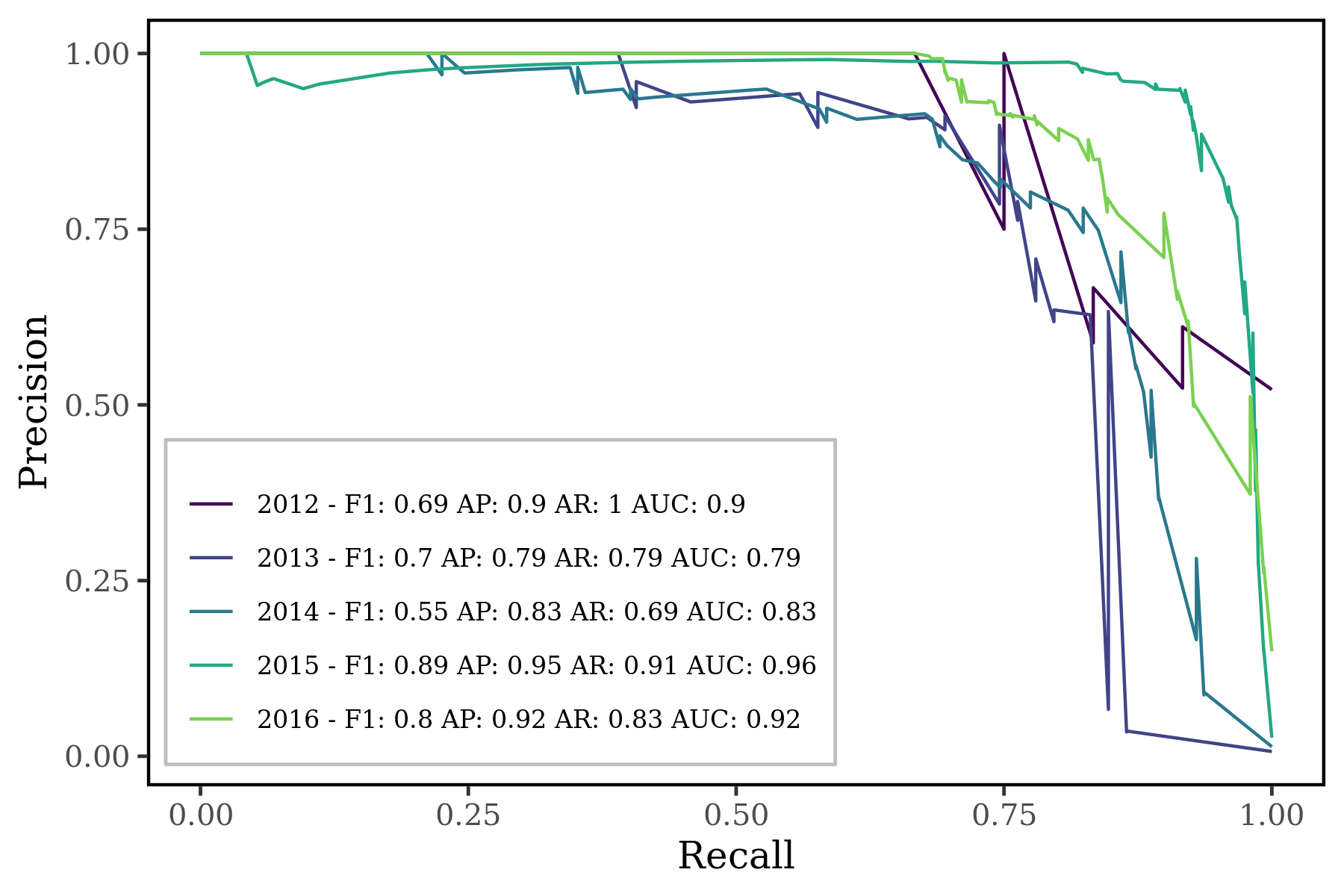}
        \caption[]{\textbf{Precision-recall Plot using Random Forest Classifier with all Variables }{\\
        The figure provides a precision-recall plot for the random forest classifier with all variables. 
        The precision-recall is used to evaluate the accuracy of the model as a function of different probability thresholds in the classifier for each prediction year. A model that is highly predictive bows towards (1, 1) in the plot. Additional statistics are provided to evaluate the performance of the classifier. An F-1 score evaluates the number of true and false positives to provide an accuracy score of 100\% if perfect. Area-under-curve (AUC) provides the integral of the precision-recall plot where a 1.0 represents a perfect prediction. Average-precision (AP) and Average-recall (AR) weights the average precision/recall across thresholds where 1.0 provides a perfect prediction.

        }}
    \label{fig:fig3}
\end{figure}

\newpage

\begin{figure}[H]
\centering
\includegraphics[width = 6in]{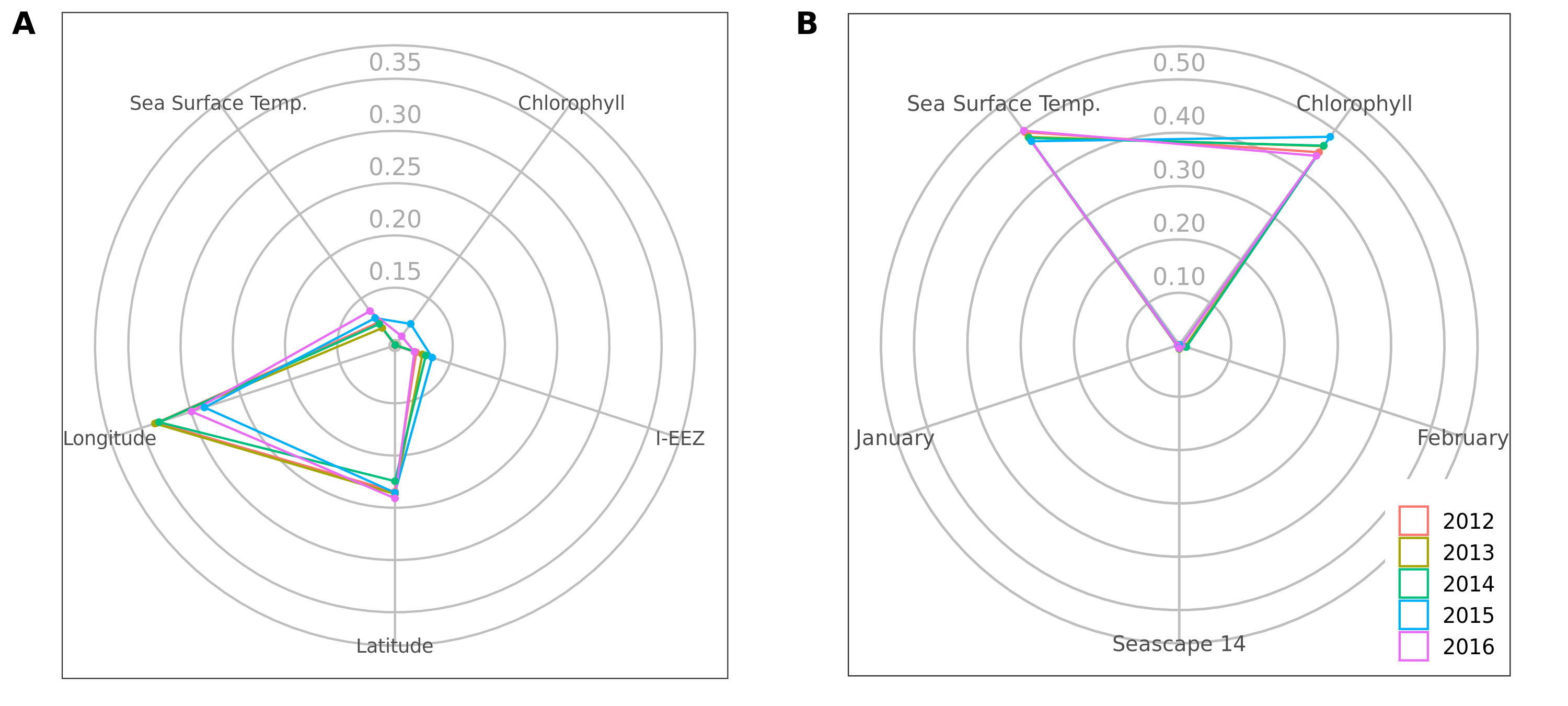}
    \caption[]{\textbf{Best Predictors from Random Forest Classifier}{ \\
    The top five predictors used for predicting illegal fishing on the Patagonia shelf from the random forest classifier for the primary model (Panel A) and model with only oceanographic variables (Panel B). In Panel A, location variables (longitude and latitude) were the most important predictors; i.e. it is an obvious condition that in order for a Chinese vessel to fish illegally in the Argentine EEZ, it must first be in the Argentine EEZ. In Panel B, sea surface temperature and chlorophyll were the best predictors.
    }}
    \label{fig:figure4}
\end{figure}

\newpage

\begin{figure}[H]
\centering
\includegraphics[]{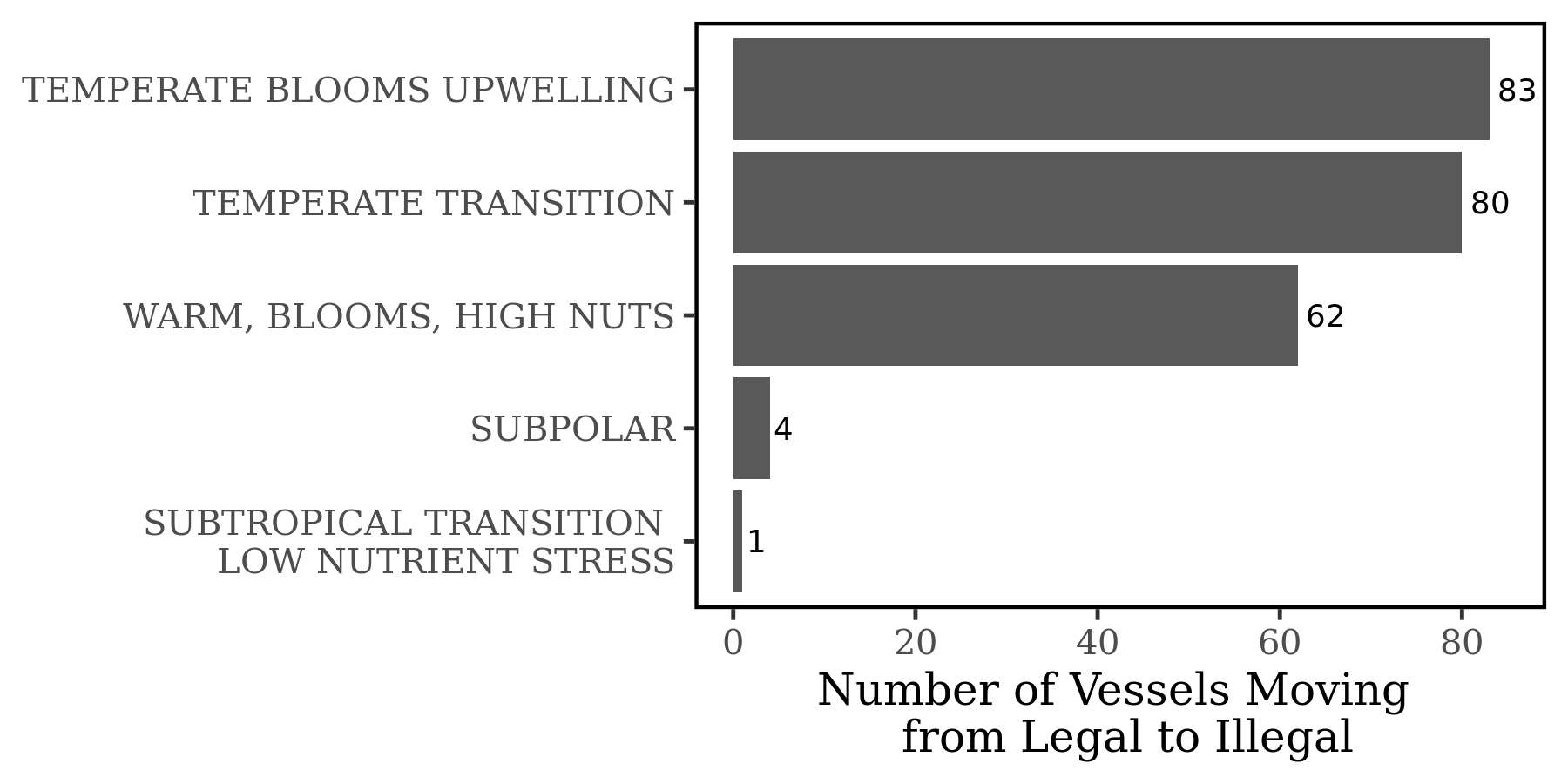}
    \caption[]{\textbf{Seascape Transition from Legal to Illegal Region}{ \\
    The number of fishing vessels that change status from fishing legally to illegal as they move to a given seascape (on the left). The behavior of fishing vessels shows they will choose to operate illegally if a seascape associated with productive fishing (e.g. the temperature bloom upwelling seascape) is within the Argentine EEZ.    
}}
    \label{fig:figure5}
\end{figure}

\newpage

\begin{figure}[H]
\centering
\includegraphics[width=6.5in]{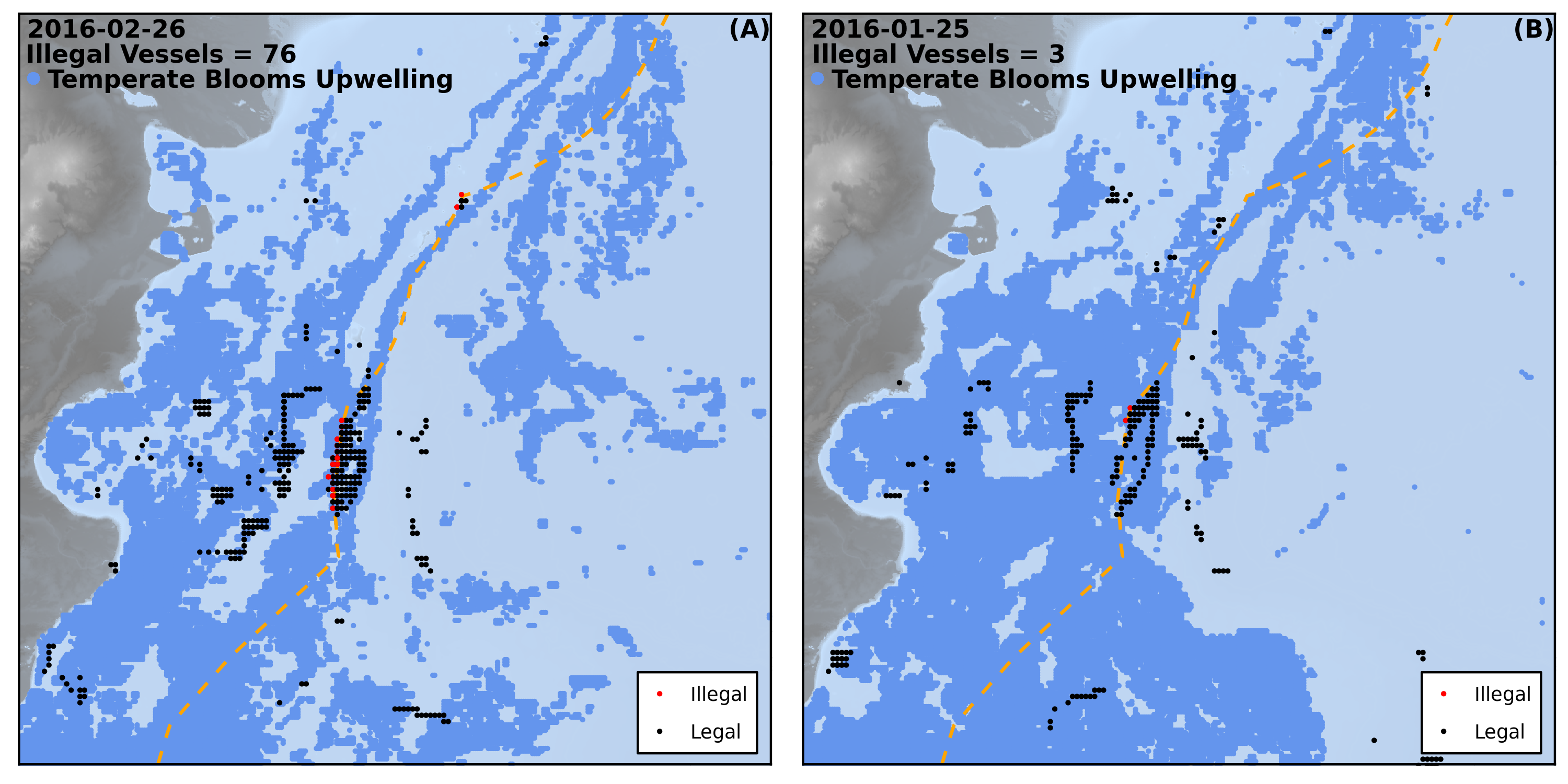}
    \caption[width = 6in]{\textbf{Seascape with Large and Small Illegal Vessel Activity}{\\
   Map of the Patagonian Shelf with high levels of illegal fishing (Panel A n=76) and low levels of illegal fishing (Panel B n=3) for two days in 2016. The EEZ is shown as an orange dashed line. Red dots represent illegal fishing vessels and black dots represent legal fishing. An important seascape associated with fishing -- the "temperate bloom upwelling" seascape is shown in dark blue.  In general, as the area covered by this seascape diminishes we see a rise in the number of illegal fishing events. This suggests that a spatial concentration of fishing effort, driven by changes in the spatial distribution of certain seascapes, can promote illegal fishing in the region.
}}
    \label{fig:figure6}
\end{figure}

\end{document}